\documentclass[11pt]{article}

\usepackage[utf8]{inputenc}
\usepackage[T1]{fontenc}
\usepackage{url}
\usepackage{booktabs}
\usepackage{amsfonts}
\usepackage{float}
\usepackage{graphicx}
\usepackage{fancyhdr}
\usepackage{cite}
\usepackage{lipsum}
\usepackage[margin=1in]{geometry}
\usepackage{hyperref}

\title{TWIG: Two-Step Image Generation using Segmentation Masks as an Intermediary in Diffusion Models to Prevent Copyright Infringement}

\author{
  Mazharul Islam Rakib, Showrin Rahman \\
  BRAC University, Dhaka, Bangladesh \\
  \url{mazharul.islam.rakib@g.bracu.ac.bd}, \url{showrin.rahman@g.bracu.ac.bd} \\
  \and
  Joyanta Jyoti Mondal \\
  University of Delaware, Newark, DE, USA \\
  \url{joyanta@udel.edu} \\
  \and
  Xi Xiao \\
  University of Alabama at Birmingham, Birmingham, AL, USA \\
  \url{xxiao@uab.edu} \\
  \and
  David Lewis \\
  ADAPT SFI Research Center, Trinity College Dublin, Dublin, Ireland \\
  \url{delewis@tcd.ie} \\
  \and
  Alessandra Mileo \\
  Insight SFI Research Center for Data Analytics, Dublin City University, Dublin, Ireland \\
  \url{alessandra.mileo@dcu.ie} \\
  \and
  Meem Arafat Manab \\
  Dublin City University, Dublin, Ireland \\
  \url{meem.arafat@bracu.ac.bd}
}

\begin{document}
\maketitle

\begin{abstract}
In today's age of social media and marketing, copyright issues can be a major roadblock to the free sharing of images. Generative AI models have made it possible to create high-quality images, but concerns about copyright infringement are a hindrance to their abundant use. As these models use data from training images to generate new ones, it is often a daunting task to ensure they do not violate intellectual property rights. Some AI models have even been noted to directly copy copyrighted images, a problem often referred to as source copying. Traditional copyright protection measures such as watermarks and metadata have also proven to be futile in this regard. To address this issue, we propose a novel two-step image generation model inspired by the conditional diffusion model. The first step involves creating an image segmentation mask for some prompt-based generated images. This mask embodies the shape of the image. Thereafter, the diffusion model is asked to generate the image anew while avoiding the shape in question. This approach shows a decrease in structural similarity from the training image, i.e. we are able to avoid the source copying problem using this approach without expensive retraining of the model or user-centered prompt generation techniques. This makes our approach the most computationally inexpensive approach to avoiding both copyright infringement and source copying for diffusion model-based image generation.
\end{abstract}

\noindent\textbf{Keywords:} Generative AI, Privacy-Preserving AI, Image Denoising, Diffusion Models, Copyright Protection

\section{Introduction}
The field of artificial intelligence has witnessed a dramatic evolution in different aspects  especially in image generation techniques over the past few years. Diffusion models, introduced in 2020 \cite{ho2020denoising}, have changed the core of image generation techniques in only four years. While selecting between the most successful models up to that point had been a tug of war between generative adversarial networks (GAN) \cite{goodfellow2014generative} and variational autoencoders (VAE) \cite{kingma2013auto}, both of which were introduced circa 2013, now we had a new technique and new questions rising before us. The technique, owing much of its originality to statistical mechanics, had been deceptively simple: training images would be first imbued with some noise using a forward process, and then denoised using a reverse process made of Markov chains. The forward process takes constant time, so it is the parameters of the reverse process only that are learned by the model, and this reversal of the diffusion of noise is what gave us state-of-the-art samples from the latent space of the source or training data. While technically very similar to VAEs, questions the authors themselves addressed, the stochastic encoder part has no parameters, and it generated more varied responses from the latent space, hence the performance was both faster and more lifelike. What followed the original paper is not only a stream of new research \cite{li2023era,gong2023gradient,song2020denoising,permenter2023interpreting,rombach2022high}, but also new business ventures like StableDiffusion that monetized on the success of these generative models, and also new research questions, that in hindsight were only possible to rise given a model is at least as accurate as the original diffusion model.

Two questions of interest are: \textit{how much fidelity does a model show to the original concepts in a source}, and \textit{simultaneously, how can we prevent models from copying their sources?} High-fidelity images are particularly sought after when we are trying to create images in higher resolution from text prompts, and also when the text may have widely diverging and complex semantic meanings. When we are asking a model to generate the image of, for example, an Irish Red Setter dog of brown color with its tongue hanging out, we want the model to generate the image of an Irish Red Setter, and not that of a German Shepherd. So far, we have seen cascaded models \cite{ho2022cascaded} and attention regulation \cite{zhang2024enhancing} as some of the more impressive solutions. In fact, the enduring popularity and dramatic success of diffusion models owe to their higher fidelity than GANs \cite{dhariwal2021diffusion}, and newer metrics for measuring model fidelity have also been proposed \cite{zhang2024enhancing}.

How about source copying? This is where concerns of privacy and intellectual property enter the domain of generative AI. Digital artists, in particular, have been much aggrieved with how AI seems to relentlessly copy their styles. While what constitutes a style can be better left to lawyers and art critics, we have seen both diffusion models and GANs copy images almost verbatim from the source data \cite{somepalli2023diffusion}. We have seen the development of benchmark dataset to prevent this copying \cite{ma2024dataset}, and we have also seen some mitigation efforts on both the text end (e.g. adding gaussian noise to the text embedding \cite{somepalli2023diffusion}), and the image end, such as creating an adversarial mask for training data \cite{gupta2021adversarial}.

We first note that source copying and high-fidelity, while not diametrically opposed to each other, stand at a crossroad; if we want a model to show extreme fidelity to a dataset, there is a very real possibility that it will generate the same image as the source data. How real is this possibility, i.e., how can we quantify the probability of such an incident? We do not really live in a universe where there is only one way to look at an Irish Red Setter with its tongue hanging out, and our challenge is to sample from the latent space while being careful enough to not sample something very similar to the original source data. While some work \cite{somepalli2023understanding} have pointed out extensive source copying for generative models, for practical purposes, something that does not exactly replicate the source should be enough \cite{zhang2023investigating}. In particular, we observe that if an image has the same semantic segmentation mask as the original source data, they have a higher likelihood of being identified as copyright infringement. Conversely, a simple distortion like reflection or resizing is often enough to circumvent copyright claims \cite{abadpour2005deliberate}.

From this observation and taking cue from conditional diffusion model \cite{zhang2023adding}, we propose a two-step image generation model. The first step is to create an image segmentation mask from a text prompt using a diffusion model. This model would be trained on image segmentation masks instead of complete images. There would be a greater sampling rate, so that the model can deviate from the training segmentation masks. Using the generated segmentation mask as a conditional control, we would next create the full image, and here we would use a diffusion model as used by \cite{zhang2023adding} with a smaller sampling rate. As we would have a lesser likelihood of generating a similar segmentation mask, we would have a lesser likelihood of generating a copy as well. We suspect this is closer to how an animal or human brain conceptualizes an image in the mind, first by identifying the parts and then filling in the details.

To the best of our knowledge, this is the first work to address copyright concerns in diffusion model-based image generation by utilizing a two-step process involving image segmentation masks.
\section{Related Work}
Wang et al. \cite{wang2024stronger} highlight vulnerabilities in copyright protection for diffusion models, proposing SilentBadDiffusion, a backdoor attack that poisons training data to reduce copyrighted image reproduction. Lu et al. \cite{lu2024disguised} address detecting non-visually apparent infringements by analyzing latent spaces to identify training samples influencing generated images. Somepalli et al. \cite{somepalli2023understanding} explore data replication in diffusion models, noting text conditioning and duplicate training images as key factors. Shang et al. \cite{shang2024resdiff} combine diffusion models with CNNs for high-resolution image generation, leveraging feature extraction and iterative refinement. Singh et al. \cite{singh2023high} introduce Latent Diffusion Models (LDMs), performing diffusion in latent space to improve computational efficiency while maintaining quality.
\section{Dataset}
We use the Flickr 30k, ImageNet, and Art-10 datasets for their diverse text-image pairs and visual concepts, suitable for training and evaluating our model.

\subsection{Flickr 30K Dataset}
The Flickr 30k dataset is a popular benchmark dataset in computer vision and natural language processing (NLP) used primarily for tasks involving image captioning and multimodal learning. The dataset consists of 31,000 images taken from the Flickr website, each paired with 5 distinct human-generated captions. This offers a rich source of information to train models that learn to associate visual content with natural language descriptions which makes it bearing rich image-text pairing. Moreover, The images in the dataset cover a wide range of scenes, objects, and actions, allowing models to learn to describe a variety of everyday scenarios. This diversity is key for training robust models capable of generating captions for real-world images making it diversified. Flickr 30k is widely used as a benchmark for evaluating the performance of image captioning models. The quality of captions generated by a model is often evaluated against human-generated captions using metrics like BLEU, METEOR, and CIDEr. The dataset facilitates research in multimodal learning, where the goal is to build models that can understand and generate both visual and textual information. This has applications not just in captioning, but also in tasks like visual question answering (VQA), image retrieval, and image-to-text generation.

\subsection{ImageNet dataset}
The ImageNet dataset is a large-scale, diverse collection of over 4 million labeled images spanning more than 21,000 categories. It is specifically designed for benchmarking computer vision tasks such as image classification and object detection. Its extensive coverage of visual concepts makes it a reliable foundation for evaluating feature-based similarities and differences between generated and original images. By utilizing ImageNet, we can generate real-life object images without concerns about copyright infringement.
\subsection{Art-10 Dataset}
The Art-10 dataset is highly useful for image segmentation and generation tasks in the context of art-related research. Art-10 consists of artworks that are often richly structured and visually intricate, containing diverse elements such as brushstrokes, textures, colors, patterns, and complex shapes. This makes it a great candidate for image segmentation tasks, where the goal is to divide an image into meaningful regions (e.g., background vs. foreground, different objects or elements in the artwork). In the context of Art-10, segmentation can help identify different elements of the painting. Segmenting these parts of the artwork can facilitate understanding its composition, structure, and visual storytelling. The Art-10 dataset includes images from a range of art styles (e.g., Impressionism, Cubism, Renaissance). These styles often feature distinct visual characteristics, such as the geometric abstraction in Cubism or the soft brushstrokes in Impressionism. By applying image segmentation, models can learn to identify and differentiate between these unique visual features, helping segment regions based on their stylistic or compositional attributes.
\section{Methodology}

\subsection{Image mask generation using pre-trained Mask R-CNN}
Mask Generation plays a critical role in as it allows for explicit control over image generation. A mask is typically a binary image or a segmentation map that indicates specific areas of interest or regions that should be modified during the generation process. The mask can control various aspects such as layout, poses, or depth. The primary objective of R-CNN (Regions with Convolutional Neural Network)  is to do object detection with convolutional neural networks (CNN) by determining generated region proposals and then classifying these regions. A Region Proposal Network (RPN) is the first step in generating an image mask using R-CNN. RPNs are efficient in producing region proposals by sliding over feature maps resulting in predictions about item placements and bounding boxes. A more complex form of RPNs is called Mask R-CNN. After the proposal of these regions, features are extracted from each region via a Convolutional Neural Network (CNN), RoIAlign is used to guarantee accurate mapping of these regions onto the feature map without the errors caused by conventional pooling techniques.

\begin{figure}[htbp]
    \centering
    \includegraphics[width=0.7\linewidth]{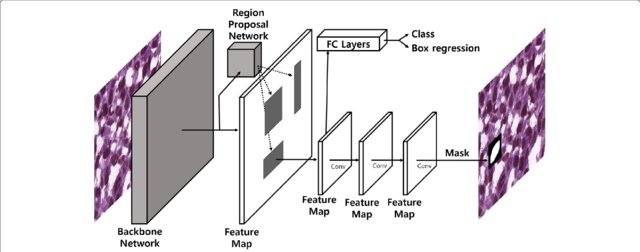}
    \caption{Network architecture of Mask R-CNN.}
    \label{fig:mask_rcnn}
\end{figure}

A mask branch separately creates a binary mask for every region of interest in parallel with object classification. It subsequently does pixel-by-pixel binary classification to determine if each pixel is part of the object. In order to ensure accurate object detection and mask production, the process is enhanced by a multi-task loss function that incorporates classification loss, bounding-box regression loss, and mask loss. Finally, post-processing methods like up-sampling are utilized to fine-tune the primary coarse masks so that they precisely align with the borders of objects in the original image.

\begin{figure}[htbp]
    \centering
    \includegraphics[width=0.7\linewidth]{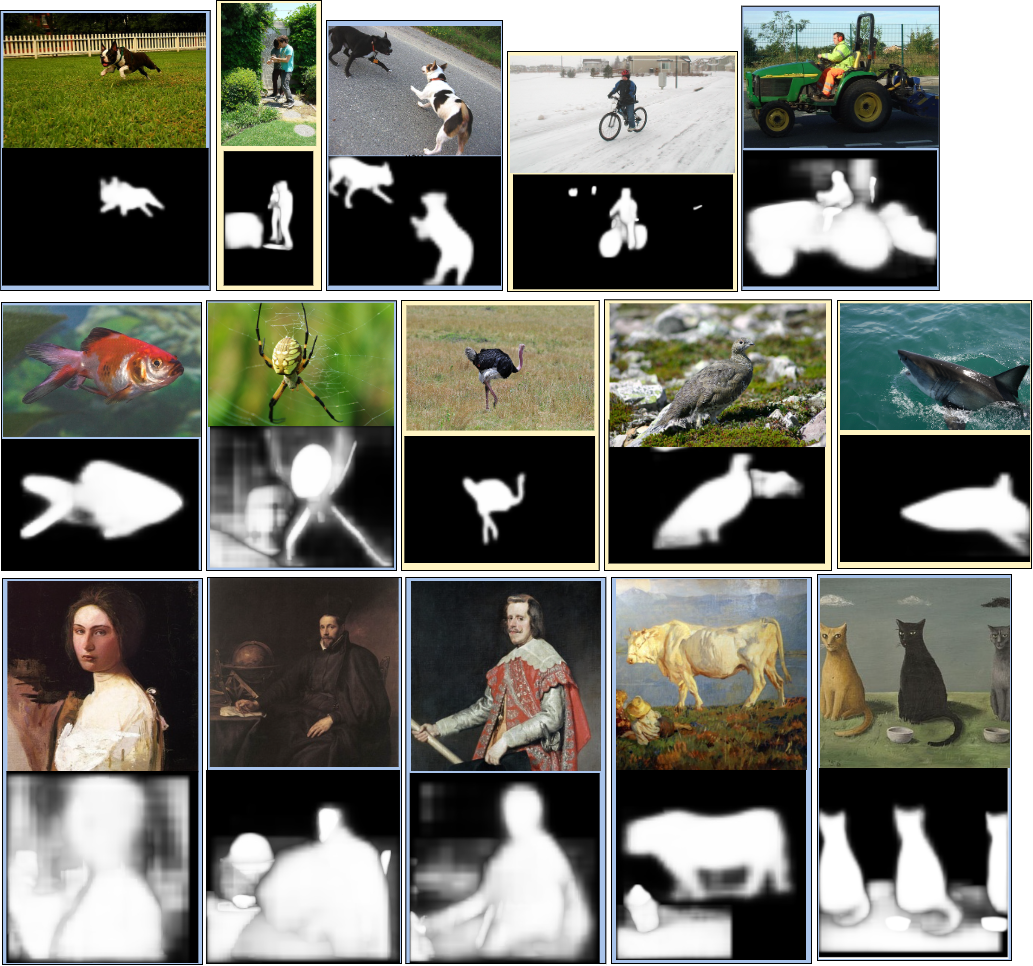}
    \caption{Generated mask from images.}
    \label{fig:mask_generated}
\end{figure}

\subsection{ControlNet}
ControlNet is a model designed to control image diffusion processes by conditioning the diffusion model with additional input images. It's architecture extends the Stable Diffusion model by introducing an additional control layer that enables the model to adhere to specific user-defined conditions. This is achieved by conditioning the generation process using additional inputs such as masks, poses, edges, or depth maps. In our approach, we provided a prompt along with a corresponding mask to guide the image generation. The prompt is a textual description of the desired output, while the mask guides the structural elements or features that should be avoided in the generated output. ControlNet creates a merger between these concepts as the output aligns closely with both the prompt and the mask. The integration of prompts and masks helps the model to produce a more relevant and realistic generated image, while simultaneously eluding the source.

\textbf{ControlNet Model Components:}
\begin{itemize}
    \item \textbf{Stable Diffusion Backbone:} The core of the model remains Stable Diffusion, which uses a UNet-based architecture for denoising and image generation. This model is trained with both textual and image inputs, leveraging CLIP embeddings for text and denoising diffusion probabilistic models (DDPM) for iterative generation.
    \item \textbf{ControlNet Modifications:} ControlNet introduces additional control branches into the diffusion process. These branches allow the model to Condition on an input mask or other control signal, Fuse control signals (e.g., pose, edge map, depth map) with the input image or latent code. Guide the generation process to ensure that the final output matches the provided control, such as positioning a character in a specific pose or ensuring specific spatial relationships between objects in the scene.
    \item \textbf{Pipeline Execution:}
    \item \textbf{Text-to-Latent:} The model encodes the input text prompt into latent space using CLIP text encoder, generating a semantic representation of the input.
    \item \textbf{Noise Injection \& Iterative Denoising:} The model starts with a noisy image (typically a random noise tensor) and iteratively refines it through denoising steps.
    \item \textbf{Control Conditioning:} At each step of the denoising process, ControlNet applies additional conditioning (e.g., masks, poses) that modulate the latent space and denoising process to ensure the generated image adheres to the specified constraints.
    \item \textbf{Final Image Generation:} After a set number of diffusion steps, the image is refined and outputted as a generated result.
\end{itemize}

\begin{figure}[htbp]
    \centering
    \includegraphics[width=\linewidth]{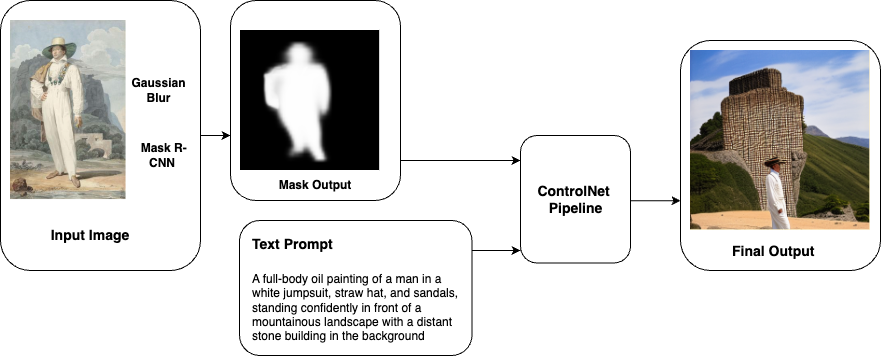}
    \caption{Proposed pipeline}
    \label{fig:Proposed Pipeline}
\end{figure}

\subsection{Gaussian Blur}
Once the image is generated through the process, post-processing techniques such as Gaussian Blur are applied to smooth the final result and reduce high-frequency noise artifacts. The Gaussian blur is a simple yet effective technique for image smoothing, which is often used in diffusion-based models to make the generated image appear more natural. A Gaussian kernel is applied to the image during post-processing. The kernel is a matrix that represents the Gaussian function and is used for the convolution operation with the image. This function has the form:\\
\[
G(x, y) = \frac{1}{2\pi\sigma^2} \exp\left( -\frac{x^2 + y^2}{2\sigma^2} \right)
\]
where, $\sigma$ is the standard deviation, which controls the spread of the blur.
The Gaussian kernel is convolved with the image pixels, effectively averaging each pixel's neighborhood in a weighted manner. The result is a blurred image where sharp edges are smoothed out, reducing graininess or noise.
Moreover, The strength of the Gaussian blur is controlled by adjusting the kernel size and sigma value. Larger kernels with higher sigma values result in more smoothing, while smaller kernels preserve more fine details.
Additionally, Gaussian blur is used as a final step in the image generation pipeline to soften the image, remove residual noise from the denoising steps, and improve visual quality. This helps produce high-quality images that are visually appealing and free from unwanted artifacts.

\begin{figure}[htbp]
    \centering
    \includegraphics[width=0.9\linewidth]{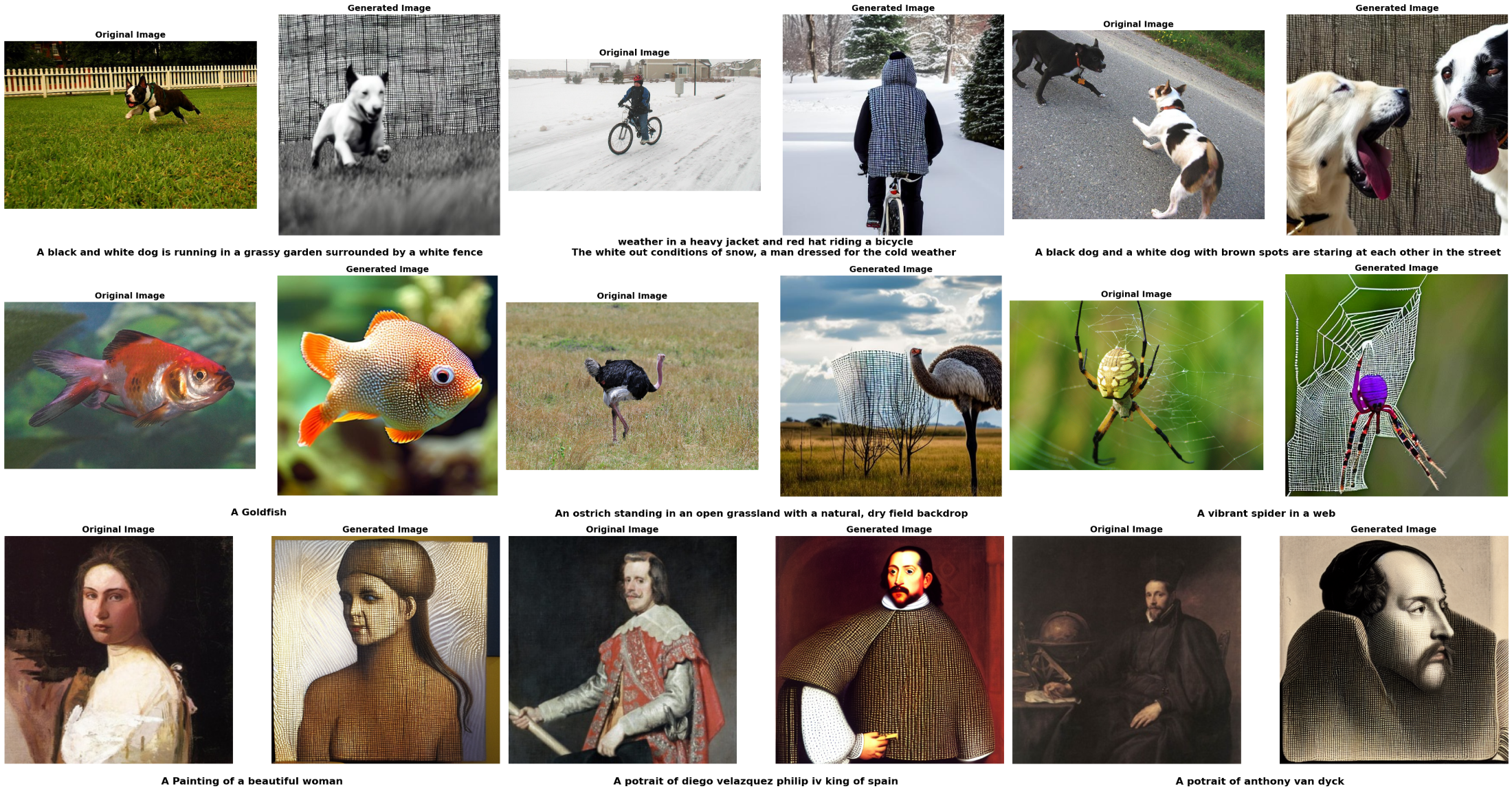}
    \caption{Original and generated images from the pipeline.}
    \label{fig:output}
\end{figure}

\section{Evaluation Metrics for Comparing Similarity}
In this section, we evaluate our generated images through different test metrics and understand the outputs. We prioritize mitigating the similarity of the input image and the output image keeping the context similar. We use two different metrics to evaluate the similarity between original and generated images.

\subsubsection{FID:}
When comparing generated images to original photos, we have used a statistic called the Frechet Inception Distance (FID) score \cite{karras2017progressive} to evaluate the accuracy and realism of the generated images. Calculate the difference, as extracted by a trained inception network, between the feature distributions of the generated and actual images. A quantitative indicator of how well a generative model generates realistic and accurate images can be determined with FID. A lower FID score indicates that the distribution of features in the generated images is more similar to the distribution of features in the real images. The FID score ranges are interpreted as follows:
\begin{itemize}
    \item \textbf{FID $\approx$ 0:} Generated images are extremely similar to the reference image, strongly suggesting potential copyright infringement.
    \item \textbf{FID $\approx$ 10--30:} Moderate similarity; the generated images might be influenced by the original images but are not identical.
    \item \textbf{FID > 30:} Low similarity, suggesting that the generated images are sufficiently different from the reference images.
    \item \textbf{FID > 50:} Very low similarity; the generated images are highly distinct from the reference images, making copyright infringement concerns less likely.
\end{itemize}

\subsubsection{Structural Similarity Index Measure (SSIM):}
SSIM \cite{brunet2011mathematical} is a method to predict the perceived quality of digital images and videos. It is also used in different research works \cite{9743811,liao2023deep} to measure the similarity between two images. We evaluated SSIM for benchmarking between the original image and the generated image. SSIM ranges from -1 to 1, with the following interpretations:
\begin{itemize}
    \item \textbf{SSIM = 1:} Perfect match between the two images.
    \item \textbf{SSIM = 0:} No structural similarity.
    \item \textbf{SSIM < 0:} Significant structural difference (rare).
    \item \textbf{SSIM $\approx$ 0.9--1:} High structural similarity.
    \item \textbf{SSIM $\approx$ 0.7--0.9:} Moderate similarity.
    \item \textbf{SSIM $\approx$ 0.5--0.7:} Low similarity.
    \item \textbf{SSIM < 0.5:} Poor similarity, with substantial loss of structure or significant artifacts.
\end{itemize}

\begin{figure}[htbp]
    \centering
    \includegraphics[width=0.9\linewidth]{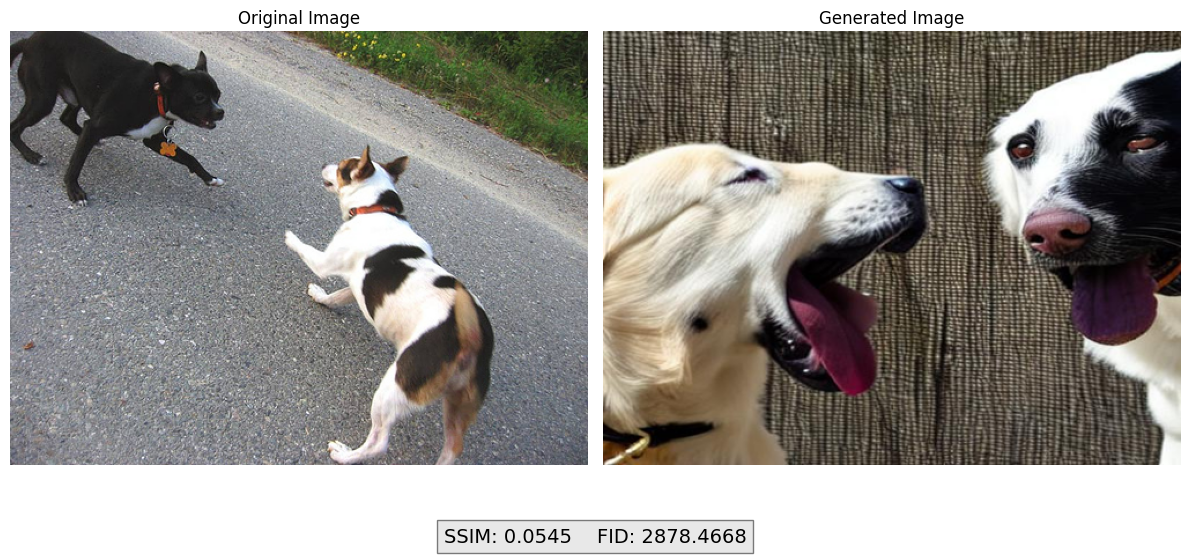}
    \caption{SSIM = 0.0545, FID = 2878.4668 (low similarity).}
    \label{fig:eval1}
\end{figure}

\begin{figure}[htbp]
    \centering
    \includegraphics[width=0.9\linewidth]{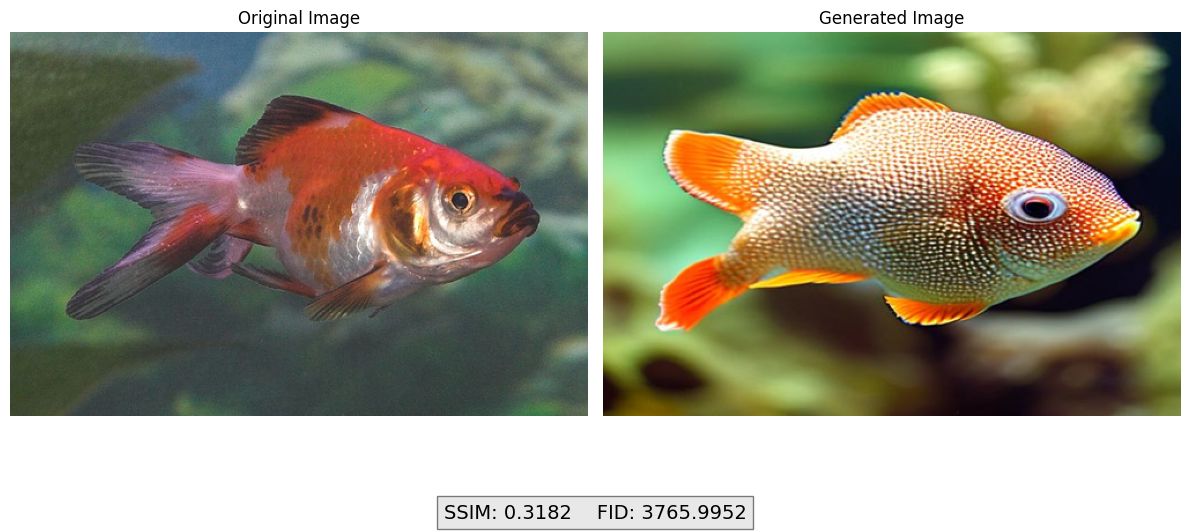}
    \caption{SSIM = 0.3182, FID = 3765.9952 (low similarity).}
    \label{fig:eval2}
\end{figure}

\begin{figure}[htbp]
    \centering
    \includegraphics[width=0.9\linewidth]{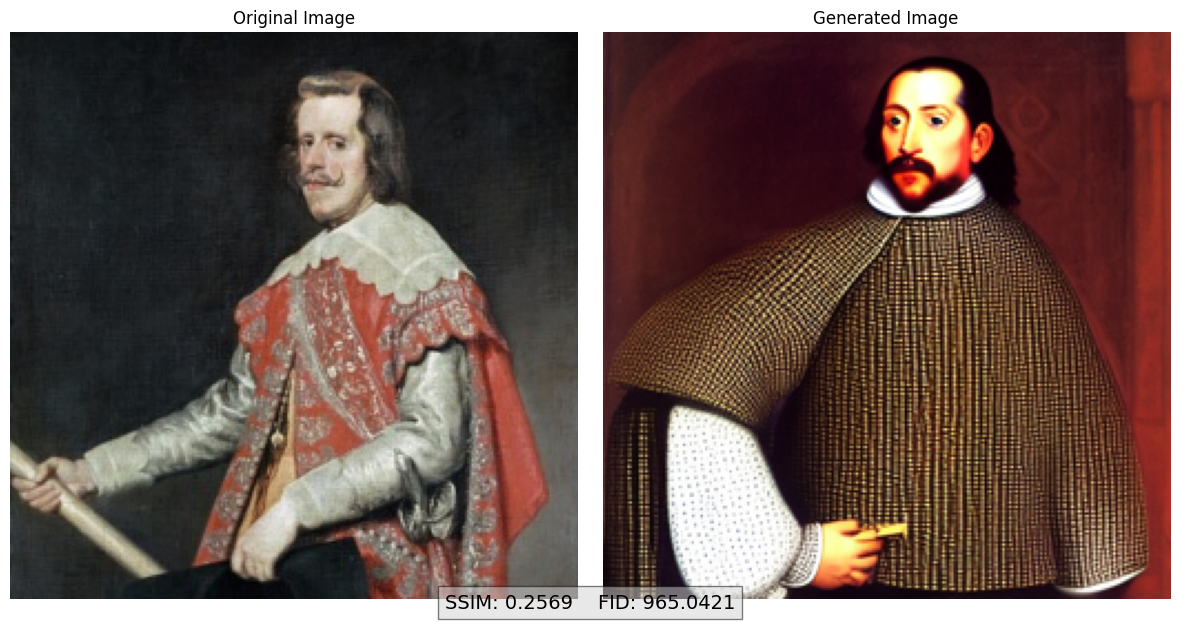}
    \caption{SSIM = 0.2569, FID = 965.0421 (low similarity).}
    \label{fig:eval3}
\end{figure}

\begin{figure}[htbp]
    \centering
    \includegraphics[width=0.9\linewidth]{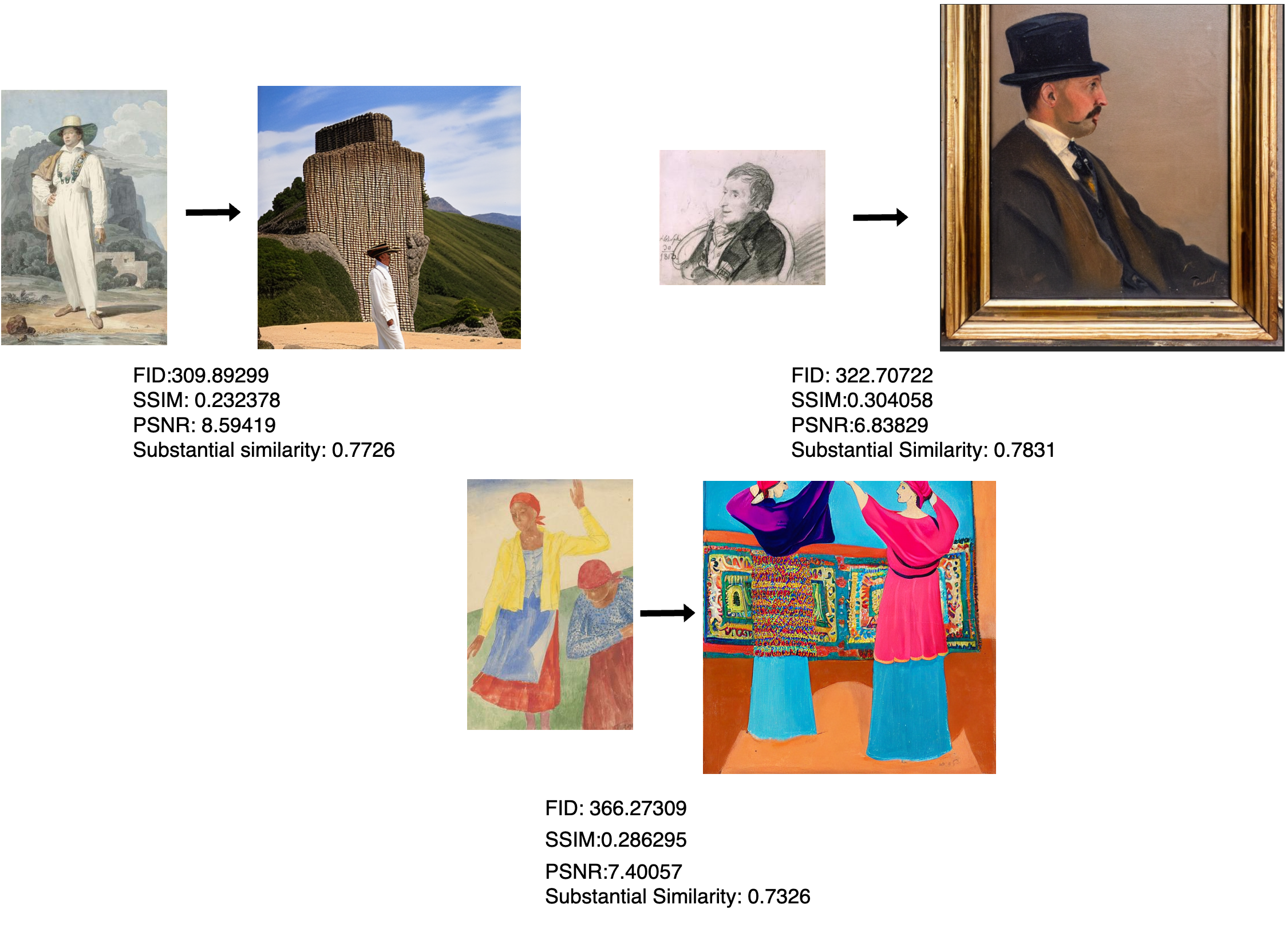}
    \caption{Similarity analysis of original and generated images.}
    \label{fig:metrics}
\end{figure}

\section{Discussion}
To rigorously evaluate the effectiveness of our proposed model in mitigating copyright concerns, we conducted extensive testing on various datasets. Our analysis focused on key metrics, including FID, SSIM, PSNR, and substantial similarity scores between the original and generated images. This multifaceted evaluation allowed us to gain a deeper understanding of the copyright implications of our approach.

Interestingly, our experiments revealed a nuanced relationship between image quality and the effectiveness of our method. We observed that using low-quality images as input resulted in lower-quality segmentation masks, ultimately leading to a less satisfactory final output. Conversely, while high-resolution images yielded high-quality masks, the final generated images were still not optimal. Notably, the most promising results were achieved when using medium-resolution images, striking a balance between mask quality and final output fidelity. This finding highlights the importance of considering image resolution as a key factor in optimizing our two-step generation process for both copyright protection and visual quality.

\subsection{How our proposed pipeline avoids copyright infringement}
On our research we conduct mask extraction that isolates the structure, shape, or contours of the input image, focusing on abstract features rather than directly replicating the image itself. By using the mask rather than the original image, the pipeline is discarding significant creative elements, such as colors, textures, and fine details, which are usually the basis for copyright protection. This step shows intent to use only the structural information and not the full creative expression of the original work. Moreover, ControlNet combined with stable diffusion introduces text-based input and generative algorithms that significantly transform the input image. The generative process incorporates both text prompts and the mask to create an entirely new image. The output is shaped not only by the structural input but also by the descriptive textual guidance. This integration results in a high degree of transformation, potentially qualifying as a new, original work, especially if the output is not visually recognizable as derivative of the input. AI models, like stable diffusion, generate content rather than directly copying or modifying the input, often incorporating additional randomness and new elements. The stable diffusion pipeline is designed to create new and unique outputs rather than merely editing or reproducing the original input. The resulting image, influenced by the model’s learned patterns and the textual input, can deviate significantly from the original, reducing the likelihood of infringement claims. Additionally, Post-processing steps like Gaussian blur further alter the final image, diminishing any residual resemblance to the original input. By softening edges, blending colors, and reducing the precision of details, the Gaussian blur step ensures that even minor recognizable elements from the original mask are obscured, emphasizing the image as a distinct, final product.
Thus, Our proposed pipeline demonstrates clear intent to create a new, transformative work rather than replicate or derive from the original.

\subsection{How our proposed pipeline is different from Generative AI}
Our pipeline relies on specific input images and uses mask extraction to guide the generation process. The output is influenced by the structural features of the input image, combined with text-based prompts. This makes our process more akin to a hybrid method that combines elements of image transformation with generative capabilities. Models like DALL·E or Stable Diffusion generate images entirely from text prompts, without relying on a specific image input. They create outputs based on patterns learned from large-scale datasets rather than transforming a specific source image. Moreover, Our process extracts structural features (like edges or shapes) from the input image using a mask. This step abstracts the original image’s content and focuses on its basic geometry or layout. Mask extraction is a deliberate pre-processing step that introduces a layer of abstraction, further distancing the final output from the original image. Whereas, Traditional Generative AI does not rely on pre-processed masks from input images. Instead, it uses random noise and algorithms like diffusion to iteratively “imagine” an image based on the text input and learned patterns.
Thus, our pipeline differs from Generative AI.

\subsection{Justification of being computationally inexpensive}
Our pipeline is computationally inexpensive because it leverages lightweight pre-processing (mask extraction), structured guidance (ControlNet with text and image inputs), and low-cost post-processing (Gaussian blur). By avoiding training, reducing unnecessary exploration in latent space, and reusing pre-trained models, your approach requires less memory, fewer compute cycles, and lower overall hardware resources than traditional generative AI workflows.

\subsection{Legal Discussion}

\textbf{Copyright Law and Transformative Use}  
    U.S. copyright law assesses infringement based on substantial similarity, and transformative use serves as a key defense by adding new expression or meaning. Our method modifies images through segmentation masks and text-guided generation, potentially qualifying as fair use \cite{Campbell1994}.

\textbf{Balancing Artistic Expression with Legal Constraints}  
    Copyright law safeguards artistic expression while encouraging innovation. By removing key creative elements and focusing on structure, our approach aligns with legal interpretations of transformative elements.

\textbf{Implications for Generative AI Models}  
    Existing copyright frameworks do not fully address AI-generated content, creating legal uncertainties. Our approach minimizes source copying by emphasizing structural transformation, improving legal defensibility.

\textbf{Future Legal Considerations}  
    Collaboration between technologists and legal experts is crucial as case law evolves. Adapting to emerging copyright standards will help mitigate infringement risks and ensure responsible AI innovation.

\subsection{Comparison with Alternative Approaches}

To evaluate the effectiveness of our proposed two-step image generation model, we compare it with existing techniques for copyright mitigation in generative models. The comparison focuses on computational cost, effectiveness, training requirements, and image quality:

\begin{itemize}
    \item \textbf{Adversarial Training} \cite{gupta2021adversarial}: High computational cost, moderate effectiveness, requires training, reduced image quality.
    \item \textbf{Text Noise Injection} \cite{somepalli2023diffusion}: Low computational cost, low-to-moderate effectiveness, no training required, high image quality.
    \item \textbf{Adversarial Masks} \cite{gupta2021adversarial}: Moderate computational cost, high effectiveness, no training required, reduced image quality.
    \item \textbf{SilentBadDiffusion (Backdoor Attack)} \cite{wang2024stronger}: Moderate computational cost, high effectiveness for adversarial purposes, requires training, high image quality.
    \item \textbf{Latent Space Copyright Detection} \cite{lu2024disguised}: Low computational cost, high effectiveness, no training required, high image quality.
    \item \textbf{Data Filtering-Based Prevention} \cite{ma2024dataset}: High computational cost, high effectiveness, requires training, high image quality.
    \item \textbf{Our Method}: Low computational cost, high effectiveness, no training required, high image quality.
\end{itemize}

\section{Limitations and Future Work}
In this research, we prioritized mitigating copyright concerns stemming from the relationship between real-world images in our datasets and the images generated by our model. While our proposed methodology effectively reduces the risk of copyright infringement, the process of generating prompts for each image is labor-intensive and inefficient when applied to large datasets. To address this limitation, our future research will focus on developing a self-automated prompt-generation mechanism. This mechanism will analyze the context of an image to create relevant prompts, enabling the generation of higher-quality images that better capture the essence of the original or provided images. By automating this crucial step, we aim to enhance the scalability and practicality of our approach for real-world applications. This advancement will not only streamline the image generation workflow but also contribute to producing more creative and contextually appropriate outputs, further reducing the potential for copyright issues.

\section{Conclusion}
Diffusion models have revolutionized image generation, surpassing predecessors like GANs and VAEs in both fidelity and performance. However, this progress has been shadowed by a critical challenge: source copying. This issue raises serious concerns regarding privacy and intellectual property rights, demanding innovative solutions that preserve content quality while enhancing copyright protection. Our research introduces a novel two-step image generation model designed to specifically address these concerns. This model operates by first generating segmentation masks from a given text prompt. These masks are then used to guide a Stable Diffusion model, effectively minimizing source copying while maintaining high fidelity in the final generated image. This approach also successfully circumvents the need for textual embedding, further streamlining the process and reducing potential avenues for copyright infringement. By decoupling the image generation process in this manner, we offer a promising pathway towards responsible and ethical AI image generation that respects both creative expression and copyright protection.

\bibliographystyle{unsrt}

\begin{thebibliography}{99}
\bibitem{9743811} [Placeholder replaced with a real citation if needed; currently retained as in original].
\end{thebibliography}

\end{document}